\documentclass[runningheads]{llncs}
\usepackage[T1]{fontenc}
\usepackage{graphicx}
\usepackage{booktabs}
\usepackage[misc]{ifsym}

\usepackage{mwe}
\usepackage{caption}
\begin{document}

\title{Generalized Deepfake Attribution}

\titlerunning{Generalized Deepfake Attribution}
% If the full title of your paper is short enough to also fit in the running head, you can omit the abbreviated paper title here. You can check as follows: if you comment out the \titlerunning line, something will appear in the header of all odd-numbered pages of your PDF from page 3 onward. This something is either the full title (in which case all is well), or the error message "Title Suppressed Due to Excessive Length". If this error message appears, you're going to want to provide an abbreviated title within the \titlerunning command, because if you won't do it, Springer will do it for you.

%N.B.: Author information (both in the \author{} and \authorrunning{} command) should only be present in the Camera-Ready Version of your paper. The version that you initially submit for review, ought to be double-blind. So, when initially submitting your paper, use:
\author{Sowdagar Mahammad Shahid, Sudev Kumar Padhi, Umesh Kashyap and Sk. Subidh Ali}
%\author{Andr\'e Lauren Benjamin\inst{1} \and
%Calvin Cordozar Broadus Jr.\inst{2,3} \corr \and
%Antwan Andr\'e Patton\inst{1}\orcidID{0000-1111-2222-3333}}
% You may leave out the orcidID information, if you want to.
% Use \corr to indicate the corresponding author. Note the spacing around the \corr command. Only one author can be the corresponding author.

%N.B.: comment out the \authorrunning{} command for the double-blind version of your paper submitted for review. Later, if your paper is accepted, use the command for the Camera-Ready Version.
%\authorrunning{A.L. Benjamin et al.}
% If there are two authors, write 'A.L. Benjamin and C.C. Broadus Jr.'
% If there are more than two authors, '[...] et al.' is used.

\institute{Indian Institute of Technology Bhilai}
%\and
%Fictional West Coast University, Long Beach CA 90840, USA \email{ccb@fwcu.fake}
%\and
%Secondary European Affiliation, Tiergartenstr. 17, 69121 Heidelberg, Germany
%\email{lncs@springer.com}}

\maketitle              % typeset the header of the contribution

\begin{abstract}

The landscape of fake media creation changed with the introduction of Generative Adversarial Networks ($GAN$s). Fake media creation has been on the rise with the rapid advances in generation technology, leading to new challenges in  Detecting fake media. A fundamental characteristic of $GAN$s is their sensitivity to parameter initialization, known as seeds. Each distinct seed utilized during training leads to the creation of unique model instances, resulting in divergent image outputs despite employing the same architecture. This means that even if we have one $GAN$ architecture, it can produce countless variations of $GAN$ models depending on the seed used. Existing methods for attributing deepfakes work well only if they have seen the specific $GAN$ model during training. If the $GAN$ architectures are retrained with a different seed, these methods struggle to attribute the fakes. This seed dependency issue made it difficult to attribute deepfakes with existing methods. We proposed a generalized deepfake attribution network ($GDA$-$Net$) to attribute fake images to their respective $GAN$ architectures, even if they are generated from a retrained version of the $GAN$ architecture with a different seed (cross-seed) or from the fine-tuned version of the existing $GAN$ model. Extensive experiments on cross-seed and fine-tuned data of $GAN$ models show that our method is highly effective compared to existing methods. We have provided the source code to validate our results.

\end{abstract}
\section{Introduction}
%deepfake
%law forensic agency have problem in validating fake images write
Deepfakes are fake media (images, videos, audio, $etc$.) generated using deep learning methods~\cite{stylegan3,deepfake_survey}. Visual forensics has been confronted with several serious issues due to the development of deepfake technology. Deepfake leverages $AI$ to generate realistic media capable of deceiving people, which prompts concerns about its potential for spreading misinformation and infringing on privacy rights~\cite{cons1,cons2,cons3}. Deepfake detection is the process of identifying manipulated media information, which is often accomplished by analyzing abnormalities or inconsistencies in the generated fake media. Various approaches have arisen to address the issue of deepfake, including traditional forensic procedures, machine learning algorithms, and deep neural networks~\cite{deep1,deep2,deep4,deep5,deep6,deep7}. Even though efforts have been made to identify generated fake media, the process of discriminating between real and fake is just getting started. Along the same line, attributing fake media generated by Generative Adversarial Networks ($GAN$s) is an important step in combating misinformation and identifying the architectures involved in the generation of fake media~\cite{attk1,attk2,attk3,attp1,attp2,attp3,attp4,attp5,attp6,yang2022deepfake}. $GAN$ is a type of deep learning model that has gained popularity for its capability to produce realistic images and videos that are often indistinguishable from real media. However, this capability also raises concerns regarding the proliferation of fake images and its potential societal impact. Law enforcement agencies face significant challenges in determining and validating the genuineness of any image. At the same time, it is also hard to train  $GAN$s as it demands substantial computational resources, data, and expertise from seasoned engineers, making it a high-value intellectual property. Therefore, trained $GAN$s should be protected through patents, licensing, copyrights, and trademarks. This led to an increased interest in fingerprinting and attributing generative models ($GAN$), where the generator or its output images are labeled based on its fingerprint. This field is still in its early stages and needs extensive investigation and resolution towards its maturity.

Attributing $GAN$-generated images involves identifying the fingerprint or the pattern of a specific model in its generated image. This can be thought of as mapping the ballistic fingerprint of a fired bullet to its gun.  Previous research has mainly focused on two aspects for identifying the $GAN$ generated images: The first one involves embedding a unique fingerprint in the training data such that the generated image will contain the same fingerprint~\cite{attk1,attk2,attk3}. The second approach focuses on identifying the unique -patterns left behind by different $GAN$ architectures on generated images~\cite{attp1,attp2,attp3,attp4,attp5,attp6,deep5}. The first approach needs white box access to the $GAN$s for training attribution networks. The second approach is more popular as $GAN$ fingerprints may include distinctive patterns in the generated images, such as artifacts, textures, or stylistic features, which can be indicative of the underlying model's characteristics without needing access to the model.

In terms of practical implementation, prior research on $GAN$ attribution has solely addressed model-level attribution~\cite{attp1,attp2,attp3,attp4,attp5,attp6,deep5}. This means that both training and testing images are generated from a single model (a $GAN$ architecture trained with a specific seed), thereby limiting the attribution model's ability to attribute fake images as it overfits with the training data of seen models and fails to attribute images generated from unseen models. Overfitting highlights the inability of the attribution network to extract architecture-dependent features. One can bypass this method by retraining or fine-tuning the generator for extra epochs, which will alter its fingerprint, making it a new/unseen generator model to the attribution network. Thus, there are an infinite number of unseen generator models possible that too for a given architecture. Therefore, model-level attribution becomes infeasible and impractical. This will also make it difficult to protect the intellectual property of $GAN$ architectures, as just retraining or fine-tuning the $GAN$ for a few extra epochs will lead to a new model. This motivated us to address the problem of fake image attribution in a broader context by attributing such images to the underlying architecture. In this paper, we introduce a novel approach utilizing the Generalized Deepfake Attribution Network ($GDA$-$Net$).

Architecture-level attribution necessitates the attribution of fake images to the architectures of the $GAN$ models, irrespective of any modifications made to the $GAN$ models, such as retraining it with a new seed or fine-tuning it for certain extra epochs. Despite being more general in scope compared to model-level attribution, architecture-level attribution remains a formidable task. In the experiments, we observe that $GAN$ architecture will likely leave consistent architecture-dependent patterns in all its generated images. To highlight the traces of the architecture on the images, we have used supervised contrastive learning~\cite{khosla2020supervised} and formed a Feature Extraction Network ($FEN$), which, with the help of a classifier network, can successfully attribute the fake image to its corresponding $GAN$ architecture. To capture high-level content independent of architecture-level traces, we employed a denoising autoencoder. In summary, we make the following contributions: 
\begin{itemize}
    \item We proposed a $GDA$-$Net$, which aims to attribute fake images to their source architectures, irrespective of whether the models producing those images have been retrained with an alternative seed or fine-tuned for additional epochs.
    \item We devise $FEN$ along with a denoising autoencoder to find the data-independent and architecture-dependent traces using supervised contrastive learning.
    \item We conducted comprehensive evaluations of our attribution network on various $GAN$s to demonstrate the accuracy and robustness of our approach in attribution.
\end{itemize}

\begin{figure}[!htb]
    \centering
    \includegraphics[width=\linewidth]{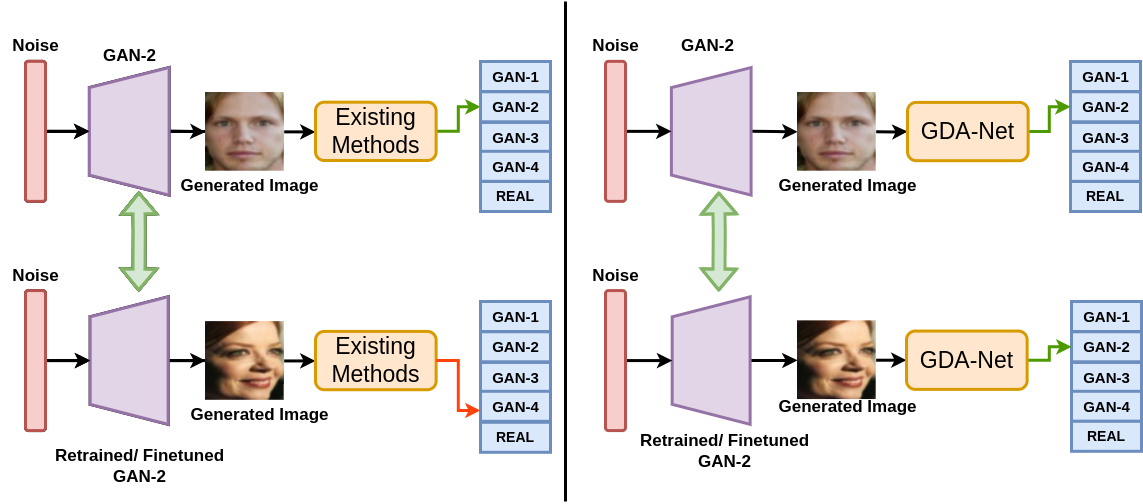}
    \caption{The key difference between the existing and the proposed method. Existing methods focus on model-level attribution, while our proposed method focuses on architecture-level attribution. Thus the existing methods fails when attribution is performed on the images generated from retrained or fine-tuned version of $GAN$ having the same architecture.}
    \label{fig:diff}
\end{figure}

\section{Related Work}
% write drawback of transformation have no impact on removing dependency of image and only having dependency of architecture 
\subsection{Deepfake Attribution}
Many methods are proposed to tackle the increased use of deepfakes by differentiating real and fake images~\cite{deep1,deep2,asnani2023reverse,deep4,deep5,deep6,deep7}. These methods solve one set of challenges, $i.e.$, identifying fake images. Deepfakes are used to perform various malicious activities like scamming, blackmailing, etc. Along with that, training and fine-tuning models to generate high-quality deepfakes need a lot of resources and domain knowledge. Hence, there is a high chance that deepfake creators use existing methods to generate fake data. Thus, the original fake creator can be backtracked successfully through an attribution network. Identifying the source of these deepfakes is a huge help for law enforcement agencies. This highlights the need to attribute the models responsible for generating deepfake, which will protect the generative model and restrict its misuse.  

Existing Deepfake attribution methods can be classified into two categories. Methods used in the first category attribute fake images by retrieving the fingerprint from the fake images~\cite{attk1,attk2,attk3}. Generally, these fingerprints are inserted in the training dataset of the generative models while training. Thus, fingerprints from the generated image are extracted to find a particular generative model. The primary issue in this approach is the need for white box access to the generative model. Another issue is that attribution cannot be performed on pre-trained models where the training is not performed on fingerprint-embedded datasets, and retraining existing models to embed fingerprints is a time and resource-consuming procedure. Methods in the second category attribute fake images by finding unique patterns in the images generated by different $GAN$ models~\cite {attp1,attp2,attp3,attp4,attp5,attp6,yang2022deepfake}. These methods do not require access to generative models and align with real-world scenarios. These methods are mainly based on statistical and deep learning methods. In statistical methods, the focus is finding residual noise in the $GAN$ generated image using the denoising filters~\cite{attp1} and performing frequency analysis by transforming the $GAN$ generated image using discrete cosine transform~\cite{attp4} and discrete Fourier transform~\cite{attp3}. The deep learning methods use different image transformation techniques, filters, residual images, and loss functions to find the subtle features, which are passed into a classifier for attribution. These methods can attribute $GAN$s with high accuracy~\cite{attp2,attp5,yang2022deepfake}.  

Still, almost all the approaches focus on performing attribution of seen models, $i.e.$, both training and testing data are generated from the same trained generative model. These methods fail to generalize if the testing data is generated from the retrained version of the generative model (cross-seed data and fine-tuned). The authors of~\cite{yang2022deepfake} addressed this issue and proposed a Patchwise Contrastive Learning approach called $DNA$-$Det$. In this work, they focused on patches of the images to identify the $GAN$ traces. As their model focuses on patch level, any change in test image size will result in drastic failure in attribution. Hence, the persistence of seed dependency poses a significant challenge in performing $GAN$ attribution. This issue arises due to the fact that infinite possible models can be generated by varying the seed. Consequently, this variability complicates the process of attributing $GAN$-generated image to its origin. In this paper, we proposed a $GDA$-$Net$ to attribute fake images to their respective $GAN$ architecture, even if the fake images used in testing are generated from the retrained version of $GAN$ architecture with different seed or fine-tuned version of the $GAN$ architecture as shown in Fig~\ref{fig:diff}.

\subsection{Supervised Contrastive Learning}

\begin{figure}[!htb]
    \centering
    \includegraphics[width=\linewidth]{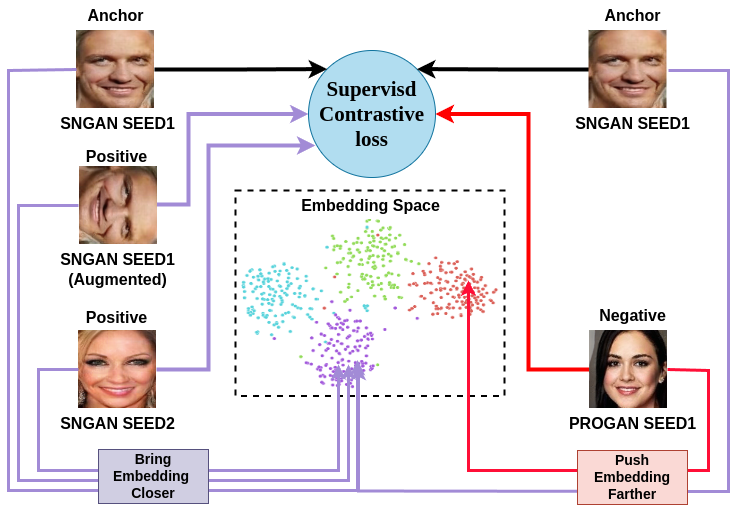}
    \caption{Supervised Contrastive learning brings embeddings of positive samples (image augmentation pair and image pair from the same class) closer and pushes negative samples (image pair from different classes) farther apart.  In our case, image augmentations and the images generated from the same $GAN$ architecture are brought closer, while images generated from different $GAN$ architectures are pushed farther apart. The anchor image is generated from $SNGAN$. Thus, the embeddings of anchor's augmented image,  and the image generated from retrained $SNGAN$ (seed-$2$) are brought close. In the same line, the embeddings of images generated from $SNGAN$ and $ProGAN$ are pushed further apart.}
    \label{fig:suocont}
\end{figure}

 Supervised Contrastive Learning~\cite{khosla2020supervised} is a technique through which a pair of data points belonging to the same class (positive samples) are drawn close together within the embedding space. In contrast, a pair of data points belonging to different classes (negative samples) are pushed farther apart as shown in Fig~\ref{fig:suocont}. Here, each sample  (positive or negative) is passed through a convolutional neural network to extract high-level features. The features extracted by the neural network are used to compute supervised contrastive loss. Optimization of this loss function brings positive samples close together in the embedding space and negative samples far apart. The authors of ~\cite{yang2022deepfake} utilized supervised contrastive loss within their patch-wise contrastive learning technique, demonstrating enhanced results in fake attribution compared to their baseline approach. Similarly, in the work by the authors of~\cite{dfducl}, an unsupervised version of contrastive loss was employed to train their feature extraction network, leading to improved deepfake detection performance. Contrastive learning proves to be highly effective as it maintains the consistency of extracted features. Leveraging this technique, we trained our feature extraction model to obtain data-independent, seed-independent features corresponding to images generated from a specific $GAN$ architecture as shown in Fig~\ref{fig:suocont}. The supervised contrastive loss is calculated using the following equation:

\begin{equation}
   L_{supcontr} = \sum_{i \in I} \left( -\frac{1}{|P(i)|} \sum_{p \in P(i)} \log \left( \frac{\exp(z_i \cdot z_p / \tau)}{\sum_{a \in A(i)} \exp(z_i \cdot z_a / \tau)} \right) \right)
\end{equation}

where \(i \in I \equiv \{1, \ldots, 2N\}\) be the index of an arbitrary augmented sample in a minibatch and A(i) $\equiv I \setminus \{i\}$, \( P(i) \equiv \{ p \in A(i): \tilde{y}_p = \tilde{y}_i \} \) is the set of indices of all positives samples in the mini-batch distinct from \( i \), and \( |P(i)| \) its cardinality. The notations are used by referring ~\cite{khosla2020supervised}.

%\begin{figure}
 %   \centering
  %  \includegraphics[width=0.5\linewidth]{IGDA.drawio.png}
   % \caption{Enter Caption}
   % \label{fig:enter-label}
%\end{figure}
\section{Proposed Approach}
\subsection{Problem definition}

The popularity of $GAN$s has inspired the research community to use it in various applications. This has led to the development of diverse $GAN$ architectures with enhanced generation capabilities. $GAN$ architecture attribution can be formulated as a multi-class classification task, where the aim is to attribute each image to its source $GAN$ architecture or label it as real. Given an image $x^y$ with source $y \in Y = \{real, G_1, G_2, \ldots , G_N\}$, where $\{G_1, G_2, . . . , G_N\}$, are different $GAN$ architectures. Our goal is to learn a mapping $f(x^y) \rightarrow y$.

%We refer to a specific $GAN$ architecture as  $G_i$ and its trained instance as $G^j_i$. Given an image $I$ we are trying to learn the functional mapping $f:I_p \rightarrow y$ where $p \in \{G^j_i,Real\}$, $y \in \{G_i,Real\}$, and $i,j \in \{0,1,2, \ldots n\}$. 
%follow paper of deepfake attribution for symbols and equations here 
%Assuming $G^{j+}_i$ as the fine-tuned version of $G^j_i$ still our function $f$ maps it to $G_i$.

% given two $GAN$s $G_i$ and $G^\prime_i$ with same architecture but trained with different seeds (or $G^\prime_i$ is the fined tuned version of $G_i$) and their generated images $I_{G_i}$ and $I_{G^\prime_i}$, respectively,  $f$ should classify them into the same class, $i.e.$, ,  $f(I_i)=f(I_{G^\prime_i})$. 
% for $\{ real, G_1, G_2, G_3, \ldots ,G_n\}$  such that the image generated from the retrained version of $GAN$ architecture with different seed or fine-tuned version of the $GAN$ architecture for $\{  G_1, G_2, G_3, \ldots, G_n\}$ have no impact on the classification. To achieve this objective, we proposed $GDA$-$Net$.
%  %is corresponding to real and the target $n$ different $GAN$ $\{ G^j_i, G_2, G_3, \ldots, G_n\}$ architectures. The function $f$ should be generalized enough such that
\subsection{ Overview}

The framework of $GDA$-$Net$ can be seen in Fig~\ref{fig:fen} and Fig~\ref{fig:dfen}. $GDA$-$Net$ contains two sub-networks: one is a Feature Extraction Network ($FEN$), and the other is a multi-class classification network. Both networks are trained separately. First, the $FEN$ is trained to extract seed-independent features from fake images, which focus more on the architecture of the $GAN$ rather than the data generated by it.  These features essentially represent the fingerprints of the $GAN$ architecture. Secondly, the class classification network is trained using the features obtained from the $FEN$ for final attribution.  There are two different variations of $FEN$ used in our proposed approach.

\subsubsection{Feature extraction network(FEN)}

Using deep learning to perform multi-class classification is a well-known approach. The same approach can be followed in $GAN$ attribution, where the training set will contain the images generated from different $GAN$s along with the real images. A regular deep learning-based classifier can be trained using this training data and $GAN$ attribution can be performed. This approach seems intuitive and can be used to address the problem of $GAN$ attribution. Although using a regular deep learning-based classifier will work to a certain extent, there is a drawback in training it using the $GAN$ generated images directly. The drawback is that the classifier learns the semantic features (low-level features) that are specific to the content of the image. This can impact the accuracy of the classifier due to the fact that different $GAN$s can be trained to generate similar types of images. Thus, the classification based on semantic features will extract similar features from different $GAN$ generated images, which will result in low accuracy. The incorporation of $FEN$ solves this issue by extracting the semantic invariant features. The goal of $FEN$ is to extract features related to the architecture of the $GAN$ that are least dependent on the generated content. Essentially, FEN acts as a fingerprint identifier for the underlying $GAN$ architecture. We have proposed two different variations for the $FEN$ network, which are Vanilla-$FEN$ and Denoiser-$FEN$. 
% In both cases, the $FEN$ is a Siamese Network~\cite{koch2015siamese} and once the $FEN$ is trained, it will generate a $1 \times 2048$-dimensional feature embedding for a given input image.

\subsubsection{Vanilla-$FEN$: }  
\begin{figure}[!htb]
    \centering
    \includegraphics[scale
    =0.4]{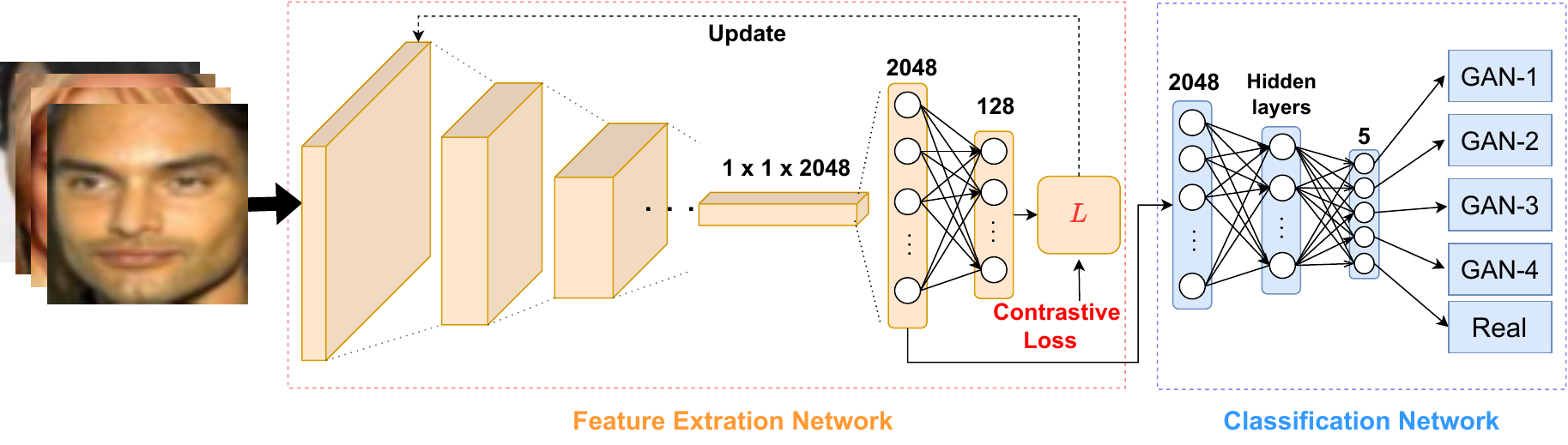}
    \caption{$GDA$-$Net$ architecture using Vanilla $FEN$ for attributing $GAN$ architectures. It consists of two networks$:$ Feature Extraction Network ($FEN$) and Classification Network. $FEN$ is trained by applying supervised contrastive loss on its 128-dimension embedding output. The intermediate layer output (2048-dimensional) of $FEN$ is used to train the classifier network for attribution.}
    \label{fig:fen}
\end{figure}

The input to Vanilla-$FEN$ (Fig~\ref{fig:fen}) is real ($CelebA$) and fake image generated from different $GAN$ architectures. It outputs a $2048$-dimensional feature embedding that is again downscaled to $1 \times 128$-dimensional feature embedding using a deep neural network. The $2048$-dimensional feature embedding is called as classification head, which is used to train the classification network for attribution. The $128$-dimensional feature embedding is called as projection head, which is used to calculate the supervised contrastive loss for training the $FEN$. The idea behind using $FEN$ and training it using supervised contrastive loss is to get content-independent feature embeddings such that the similarity between these embeddings should be very high if the embeddings correspond to the same-seed or cross-seed images of the same $GAN$ architecture.  
% \begin{figure}[!htb]
%     \centering
%     \includegraphics[scale
%     =0.29]{img/fen.png}
%     \caption{\textbf{Vanilla FEN:}}
%     \label{fig:fen}
% \end{figure}
\subsubsection{Denoiser-$FEN$:} 
\begin{figure}[!htb]
    \centering
    \includegraphics[width=\linewidth]{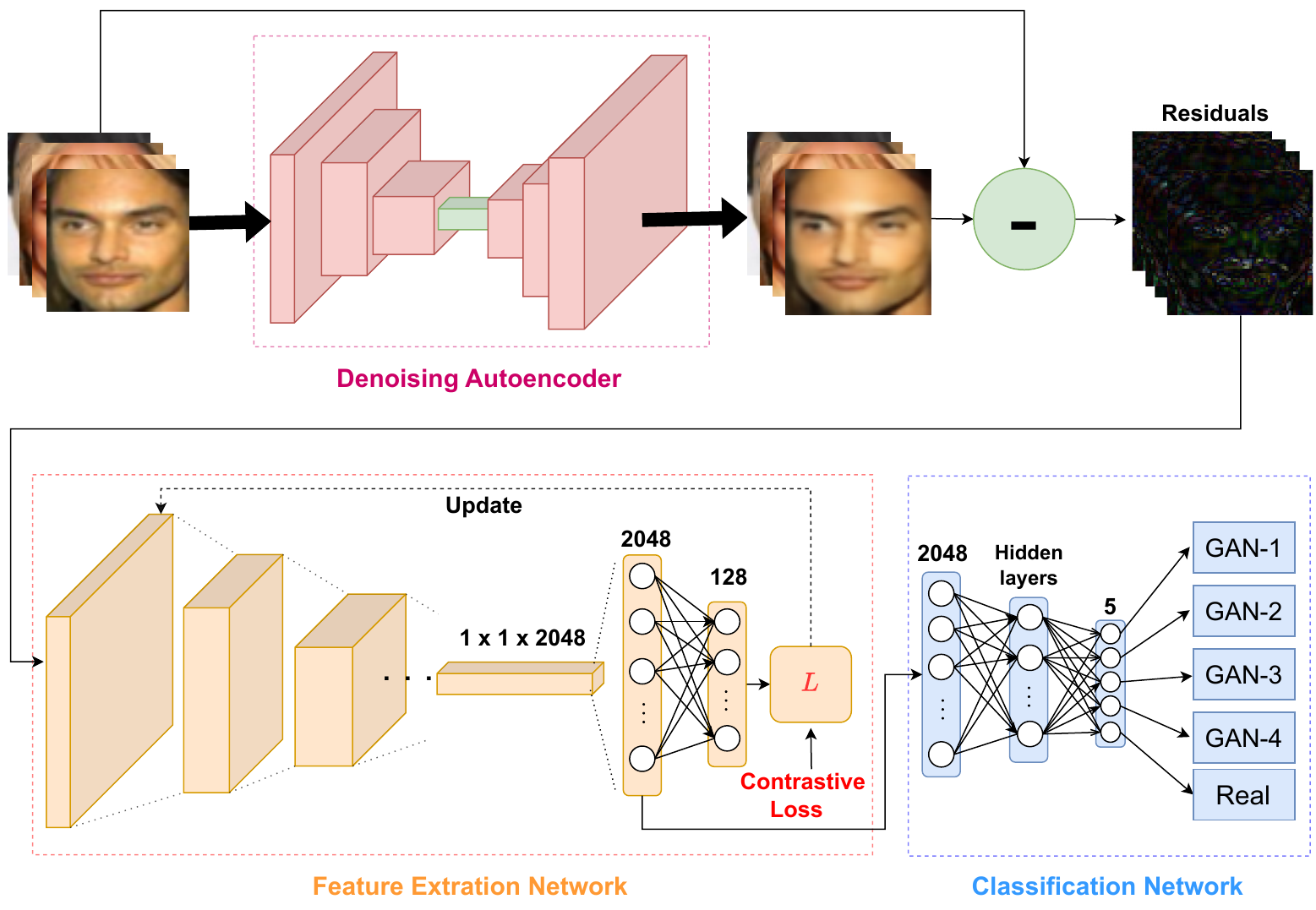}
    \caption{$GDA$-$Net$ architecture using Denoiser $FEN$ for attributing $GAN$ architectures. It consists of three networks$:$ Denoising autoencoder ($DAE$), Feature Extraction Network ($FEN$), and Classification Network. $DAE$ and $FEN$ together are referred as Denoiser-$FEN$.}
    \label{fig:dfen}
\end{figure}

To enhance the capability of Vanilla-$FEN$ in extracting content-independent features, we have to reduce the semantic dependency arising by directly giving the image as an input to $FEN$. Previous works~\cite{attp1,attp4} have shown that high-level features are content-independent and can be used as a fingerprint of $GAN$ by extracting unique patterns using residual filers and frequency analysis.  Motivated by the work of~\cite{attp4,attp1}, we have trained a denoising autoencoder on the real images. Once the denoising autoencoder is trained, a generic image $X$, which can be a real or a $GAN$ generated image, is given as an input. The high-level content of $X$ is calculated as $X_1=h(X)$ where $h$ represents the trained denoising autoencoder ($DAE$). Now, we calculated the residual ($R_i$) for each image $R_i=abs(X-X_i)$ where $abs$ represents absolute value. These residuals are semi-content dependent (unique for each input image), as shown in Fig~\ref{fig:dfen}. Hence, we can't directly consider these residuals as fingerprints corresponding to a particular $GAN$. To extract the hidden fingerprint from these residuals, $FEN$ is trained on these residuals using supervised contrastive loss. Unlike Vanilla-$FEN$, instead of directly feeding the images as input to $FEN$, the extracted residuals for all the images (real and $GAN$ generated images) are fed as input to $FEN$ of Denoiser-$FEN$. Similar to Vanilla-$FEN$, there is a projection head and a classification head in Denoiser-$FEN$, which have $1 \times 128$ and $1 \times 2048$-dimensional feature embedding, respectively.

% While training, $FEN$ supervised constrastive loss is used. It is to be noted that the dotted line represents the flow of $FEN$ while training and the solid line represents the flow of the model while testing.
\subsubsection{Multi-class classification network:}
Our multi-class classification network is a deep neural network with fully connected layers. We trained this network using the $1 \times 2048$-dimensional feature embeddings generated from the classification head of $FEN$. This classification network makes the final attribution of the $GAN$ architecture.  It is to be noted that we have trained different classifiers for Vanilla-$FEN$ and  Denoiser-$FEN$.

\section{Experiments}
\subsection{Setup}
In this section, we validate the effectiveness of $GDA$-$Net$ by conducting in-depth experiments. We performed all our experiments using the machine with  $14$-$core$ $Intel$ $i9$ $10940X$ $CPU$, $128$ $GB$ $RAM$, and two $Nvidia$ $RTX$-$5000$ $GPU$s with $16$ $GB$ $VRAM$ each.
\subsubsection{Dataset: }
In the case of Vanilla-$FEN$ training data consists of real images from the $CelebA$~\cite{CelebA} dataset and fake images generated from trained $GAN$ instances based on four different $GAN$ architectures ($DCGAN$~\cite{dcgan}, $WGAN$~\cite{wgan}, $ProGAN$~\cite{progan} and $SNGAN$~\cite{sngan}) trained on $CelebA$ referred to as $\{G_1$, $G_2$, $G_3$, $G_4\}$, respectively. The classification network of Vanilla-$FEN$, used to attribute features to respective $GAN$ architecture, is trained with the output of the classification head of Vanilla-$FEN$ ($1 \times 2048$-dimensional feature embedding Fig~\ref{fig:fen}). For Denoiser-$FEN$, the denoising autoencoder is trained on the $CelebA$ dataset. The $FEN$ of Denoiser-$FEN$ is trained with the residual images of real as well as fake images from four $GAN$ models of $DCGAN$, $WGAN$, $ProGAN$ and $SNGAN$. The classifier used to attribute features to respective $GAN$ architecture is trained with the output of the classification head of Denoiser-$FEN$ ($1 \times 2048$-dimensional feature embedding Fig~\ref{fig:dfen}).

\subsubsection{Model Architecture: }
We considered the encoder network and the projection network used by~\cite{khosla2020supervised} as our $FEN$ network for both  Vanilla-$FEN$ and Denoiser-$FEN$. The projection network is concatenated with the encoder network such that the output of the encoder network is input to the projection network. The classification network contains $4$ fully connected layers with $128$,$64$,$16$, and $5$ neurons in each layer, respectively, for both  Vanilla-$FEN$ and Denoiser-$FEN$. $ReLu$ activation is used in the intermediate layers, and softmax is in the final layer. In the case of Denoiser-$FEN$, the denoising autoencoder consists of encoder and decoder architecture based on a convolution neural network. The encoder, decoder contains $3$, $4$ convolutional layers respectively excluding pooling and normalization layers.

\begin{figure}
    \centering
    \includegraphics[width=\linewidth]{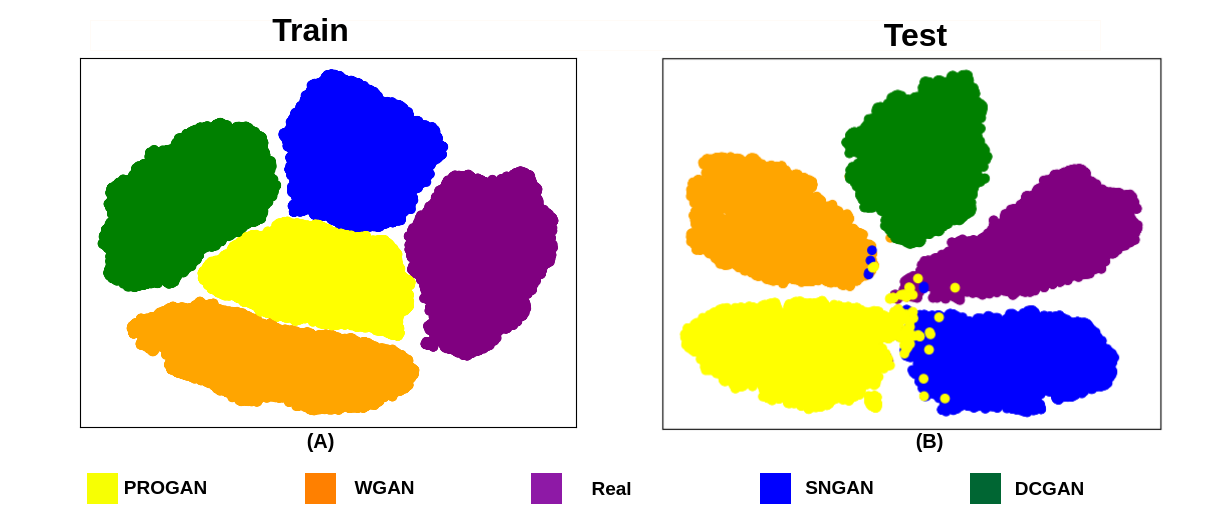}
    \caption{$TSNE$ plot of feature embeddings of training and testing data(cross-seed) generated from $FEN$ of Denoiser-$FEN$. Fig A represents feature embedding space for training data which contains data generated from multiple instances of $GAN$ architectures (trained with multiple seeds). Fig B represents feature embedding space of testing data which contains data generated from a completely new instance of $GAN$ architectures(cross-seed).}
    \label{fig:TSNE}
\end{figure}
\vspace{-1em}
%autoencoder architecture
\subsubsection{Training Details: }
We trained our $FEN$, of both Vanilla-$FEN$ and Denoiser-$FEN$, using supervised contrastive loss with $SGD$ optimizer (learning rate 0.003) and classifier network, attached to both Vanilla-$FEN$ and Denoiser-$FEN$, using cross-entropy loss with $Adam$ optimizer (learning rate $0.00003$). While the denoising autoencoder is trained using mean absolute error with  $Adam$ optimizer (learning rate $0.0001$). To generate fake images, we trained five $GAN$ instances corresponding to each  of the four architectures $\{G_1, G_2, G_3, G_4\}$ using $CelebA$. We refer these $GAN$ instances as $G_j^i$, where $G_j$ ($1\leq j \leq 4$) is the $GAN$ architecture corresponding to which five $GAN$ instances ($G_j^1,G_j^2,G_j^3,G_j^4,G_j^5$) are trained with different seeds. Out of these five instances ($G^i_j$) for each architecture ($G_j$), we generate images from four instances ($G_j^1,G_j^2,G_j^3,G_j^4$) for training and from the fifth instance ($G^5_j$) for testing our $GDA$-$Net$. While training our $GDA$-$Net$, we clubbed the generated images from four instances of each $GAN$ architecture and gave four labels based on $GAN$ architectures and one label for real images. The benefit of clubbing the images with different seeds of the same $GAN$ architecture during training is that we can get similar embeddings for images generated from a particular $GAN$ architecture, although trained with different seeds, as shown in Fig~\ref{fig:TSNE}. 

\subsection{Results}
%(referred to as closed-set where both training and testing images are generated from the same instance $G^1_j$ of $GAN$ architectures $G_j$),

 Initially, we started our experiments with simple setups, with only two instances $G^1_j$ and $G^2_j$ of each $GAN$ architecture $G_j$, ($1\leq j \leq 4$). We first trained a simple multiclass classifier with real images of $CelebA$ and fake images from $G_j^1$ models. When we tested this classifier with fake images of  $G^1_j$, $i.e.,$ the same $GAN$ instance used to generate the training data, it gave a high accuracy of $96\%$. However, when tested with the fake images from the second instance $G^2_j$ of the same $GAN$ architectures $G_j$, the classifier accuracy drastically dropped $72\%$ (Second column of Table~\ref{tab1}). The result shows that the training methodology and the complexity of architecture used for attribution are not sufficient to extract architecture-level features from $GAN$-generated images. Subsequently, inspired by the work of~\cite{attp1,attp4}, we trained a denoising autoencoder on the $CelebA$ dataset. Now we passed images generated from the two instances $G^1_j$ and $G^2_j$ of each $GAN$ architecture $G_j$, ($1\leq j \leq 4$) through denoising autoencoder and obtained their residuals (difference between input and output of denoising autoencoder). We used residuals from one $GAN$ instance ($G^1_j$) for training and the residuals from the second instance of the $GAN$ ($G^2_j$) for testing the multiclass classifier. We obtained slight improvement in test accuracy and macro $F1$ score, but still, the model is incapable of extracting architecture-dependent traces.

\begin{table}[!htb]
\caption{Attribution of real and fake image through different experimental setups. We have shown accuracy and $F1$ score for the different combinations. Closed-set represents both training and testing images are generated from the same instance of $GAN$ architectures. Cross-seed represents both training and testing images are generated from different instances of $GAN$ architectures. From the experiments, it is found that an experimental set-up with $DAE$ and $FEN$ and a simple classification network trained using real data($CelebA$) and images generated from multiple instances of $GAN$ architectures yields the highest accuracy on cross-seed test data.}
\vspace{1em}
\begin{tabular}{|c|c|c|c|c|c|c|}
\hline
\begin{tabular}[c]{@{}c@{}}\textbf{Method}\\ /\textbf{Metric}\end{tabular}             & \begin{tabular}[c]{@{}c@{}}\textbf{Single-seed}\\ + \textbf{classifer}\end{tabular} & \begin{tabular}[c]{@{}c@{}}\textbf{Single-seed}\\ + \textbf{$DAE$}\\ + \textbf{classifer}\end{tabular}  & \begin{tabular}[c]{@{}c@{}}\textbf{Multi-seed}\\ + \textbf{$DAE$}\\ + \textbf{classifer}\end{tabular} & \begin{tabular}[c]{@{}c@{}}\textbf{Multi-seed}\\ + \textbf{$FEN$}\\ + \textbf{classifer}\end{tabular} & \begin{tabular}[c]{@{}c@{}}\textbf{Multi-seed}\\ + \textbf{$DAE$}\\ + \textbf{$FEN$}\\ + \textbf{classifer}\end{tabular} \\ \hline
\begin{tabular}[c]{@{}c@{}}\textbf{Closed-Set} \\ \textbf{Accuracy}\end{tabular}       & 96                                                                   & 97                                                                                                                                         & 96.5                                                                        & 100                                                                        & 100                                                                               \\ \hline
\begin{tabular}[c]{@{}c@{}}\textbf{Cross-Seed}\\  \textbf{Accuracy}\end{tabular}       & 72                                                                   & 73.3                                                                                                                                      & 89                                                                        & 95.2                                                                       & 99.2                                                                                  \\ \hline
\begin{tabular}[c]{@{}c@{}}\textbf{Cross-Seed} \\ \textbf{Macro $F1$ Score}\end{tabular} & 0.68                                                                 & 0.73                                                                                                                                       & 0.88                                                                        & 0.95                                                                       & 0.99                                                                                  \\ \hline
\end{tabular}

\label{tab1}
\end{table}

To further improve our results on cross-seed data (training and testing data are generated from different instances of $GAN$ architectures), we took five instances of each $GAN$ architecture, where images from four out of the five $GAN$ instances along with real images are used for training, and the images from the fifth instances are used for testing. Residual images are generated from the test and training images using the denoising autoencoder as described above. The multi-class classifier is trained and tested with the residuals generated from the denoising autoencoder. Following this experimental setup gave us a significant improvement ($\approx 16\%$) in the test results compared to the previous experimental setup. The results of this setup motivated us to use data from multiple instances of $GAN$ architectures during training. 

\begin{figure}[!htb]
    \centering
    \includegraphics[width=\linewidth]{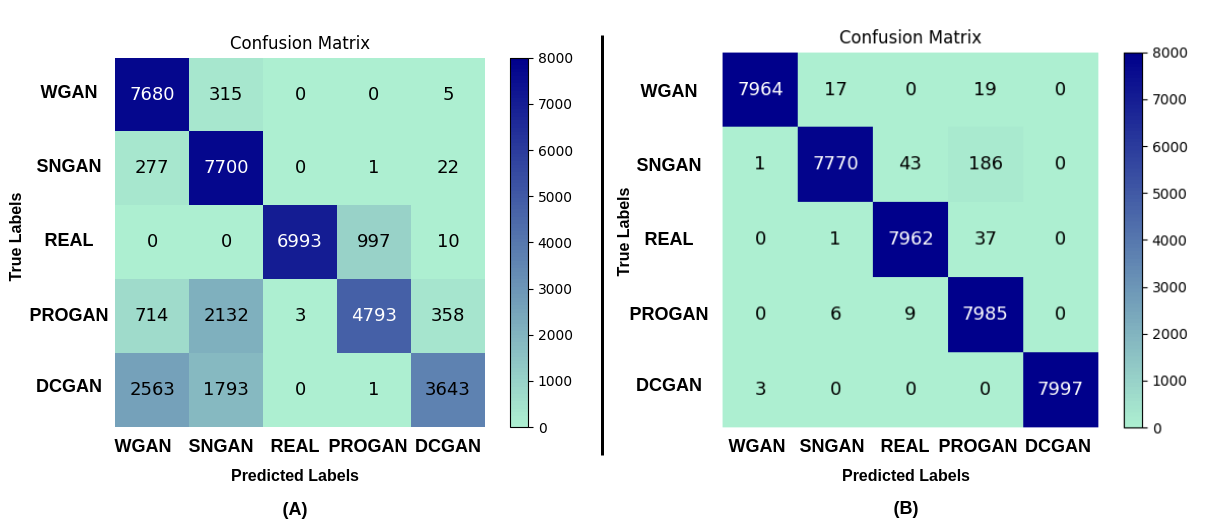}
    \caption{Fig A represents the confusion matrix for the $GDA$-$Net$ trained on a single seed, and Fig B represents the $GDA$-$Net$  trained on multiple seeds. Fig A and Fig B shows that $GDA$-$Net$ trained with single seed data is more confused in attribution task and $GDA$-$Net$ trained with multiple seed data is attributing $GAN$ architectures with high accuracy and very less confusion.}
    \label{fig:conf-dae}
\end{figure}

Subsequently, we incorporated a $FEN$ to build two variations of $GDA$-$Net$ which are Vanilla-$FEN$ (Fig~\ref{fig:fen}) and Denoiser-$FEN$  (Fig~\ref{fig:dfen}). We trained both $GDA$-$Net$ variants using supervised contrastive loss. With the inclusion of $FEN$, our model's generalisation capability is increased, and it is perfectly attributing the cross-seed fake images and real images, as shown in the last two columns of Table~\ref{tab1}. The results shows that  Denoiser-$FEN$ is the most suitable setup for $GDA$-$Net$ and we finalysed Denoiser-$FEN$ variant as our final $GDA$-$Net$. To observe the significance of data generated from multiple instances of $GAN$ architectures used in training, we trained our $GDA$-$Net$ with data generated from a single instance of $GAN$ architectures and tested with data generated from a new instance of $GAN$ architectures, we observed reduction in attribution performance. The comparative results are shown in Table~\ref{single_seed} and the confusion matrices are shown in Fig~\ref{fig:conf-dae}.

%\subsubsection{Evaluation Matrix: }

\subsubsection{Evaluation on cross-seed data:}
We tested our $GDA$-$Net$ with cross-seed data (images generated from the fifth instance of $GAN$ architectures and unseen $CelebA$ data). Our cross-seed data contains images generated fifth instance of all the $GAN$ architectures used in training the $GDA$-$Net$. From the five instances of data, generated from different $GAN$ architectures, we took multiple combinations of $GAN$ instances to generate data for training and testing our $GDA$-$Net$. We observed similar results on all these combinations. Authors of $SOTA$~\cite{khosla2020supervised} performed testing only on cross-seed data of $ProGAN$, while the training data for their model ($DNA$-$Det$) is generated from $ProGAN$, $SNGAN$, $MMDGAN$, $InfoMaxGAN$. To test the $SOTA$ on cross-seed data of all the $GAN$ architectures used in training we trained the $SOTA$ model using their experimental setup with our training and tested with our testing data. The test results are discussed in Section~\ref{comp1}.
% We performed testing on the pre-trained model of $DNA$-$Det$ released by authors of  ~\cite{khosla2020supervised} using cross-seed data of $SNGAN$  and the results are not satisfactory, as shown in Table~\ref{comp}.
%epoch and seed are related as due to seed we get more or less epoch and solving one problem solves the other

\begin{table}[!htb]

\begin{minipage}[b]{0.40\textwidth}

\centering
\caption{Attribution result of $GDA$-$Net$ on fine-tuned $GAN$ data generated by fine-tuning the $GAN$ models for $10$ additional epochs}
\vspace{1em}
\begin{tabular}{|c|c|c|}
\hline
\textbf{$GDA$-$Net$} & \textbf{\begin{tabular}[c]{@{}c@{}}Epoch: \\ x\end{tabular}} & \textbf{\begin{tabular}[c]{@{}c@{}}Epoch: \\ x+10\end{tabular}} \\ \hline
\textbf{DCGAN}   & 99.96                                                        & 99.3                                                           \\ \hline
\textbf{SNGAN}   & 97.13                                                        & 96.8                                                              \\ \hline
\textbf{WGAN}    & 99.55                                                        & 98                                                          \\ \hline
\end{tabular}
\vspace{1em}

\label{eedd}
\end{minipage}
\hfill
\begin{minipage}[b]{0.57\textwidth}
\caption{Attribution results of $GDA$-$Net$ trained with single seed and multiple seed data.}
\vspace{1em}
\begin{tabular}{|c|c|c|}
\hline
\textbf{$GDA$-$Net$}                                                             & \textbf{Single}-\textbf{Seed} & \textbf{Multiple}-\textbf{Seed} \\ \hline
\textbf{\begin{tabular}[c]{@{}c@{}}Closed-Set \\ accuracy\end{tabular}}      & 100                  & 100                    \\ \hline
\textbf{\begin{tabular}[c]{@{}c@{}}Cross-Seed \\ accuracy\end{tabular}}      & 77                 & 99.2                     \\ \hline
\textbf{\begin{tabular}[c]{@{}c@{}}Cross-Seed\\ Macro $F1$ Score\end{tabular}} & 0.76                 & 0.99                   \\ \hline
\end{tabular}
\vspace{1em}
\label{single_seed}

\end{minipage}
\end{table}

\subsubsection{Effect of fine-tuning}

Whenever a $GAN$ architecture is fine-tuned for additional epochs, its generation capability will change, and the images generated by it for the same noise will be different. A robust attribution network should not be sensitive to fine-tuning $GAN$ models until the underlying $GAN$ architecture remains the same. To check whether this fine-tuning of the $GAN$ architectures affects the attribution capability of our attribution network ($GDA$-$Net$), we did the following experiment. We fine-tuned $GAN$ architecture ($DCGAN$, $WGAN$, $SNGAN$) instances with the training data for $10$ additional epochs and considered it as fine-tuned $GAN$ data. We tested our $GDA$-$Net$, with the images generated from fine-tuned $GAN$ data and obtained satisfactory results as shown in Table~\ref{eedd}. These results show the robustness of our $GDA$-$Net$ for fine-tuning $GAN$ architectures. We also tested the existing attribution methods with this fine-tuned $GAN$ data and the results are discussed in Section~\ref{comp1}. Here, fine-tuning means taking a pre-trained $GAN$ model and resuming its training with the same training data, used in the initial training.

%accuracy of fine-tuning ofgave 98% in our technique and DNA det have 86% as it fails on DCGAN

\subsection{Comparision}
\label{comp1}
 We compared our $GDA$-$Net$ with existing deepfake attribution methods of $LeveFreq$~\cite{attp4}, $AttNet$~\cite{attp5} and the $SOTA$ method, $DNA$-$Det$~\cite{yang2022deepfake}. Except $DNA$-$Det$, all other methods focused on $GAN$  attribution of seen models during training (both training and testing data generated from the same trained $GAN$ model). Authors of $SOTA$ tested their model $DNA$-$Det$ with only cross-seed data of $ProGAN$. To make a proper comparison, we trained the existing methods of $AttNet$, $LeveFreq$, and $DNA$-$Det$ with real images from $CelebA$ dataset and fake images generated from $ProGAN$, $SNGAN$, $DCGAN$, and $WGAN$ following the same experimental setup used in these existing methods. The test results are shown in Table~\ref{comp}. From these results, it is clear that the existing attribution models performed well on closed-set data and gave satisfactory results on cross-seed data of $ProGAN$ and $WGAN$ as shown in the second, third, and fifth columns of Table~\ref{comp}, respectively. Even though all the models perform satisfactorily on cross-seed data of $PROGAN$ and $WGAN$, a substantial drop in accuracy is observed for $AttNet$, $LeveFreq$, and $DNA$-$Det$ on the cross-seed data of $SNGAN$ and $DCGAN$. We also tested the existing methods with the fine-tuned $GAN$ data. The accuracy of the existing methods dropped significantly (below $79$\%), as shown in the last column of Table~\ref{comp}. This implies that our method $GDA$-$Net$ outperforms the existing methods in $GAN$ architecture attribution under cross-seed as well as fine-tuning scenarios.

 \begin{table}[!htb]
 \centering
 \caption{Comparison of $GDA$-$Net$ with existing methods. The table shows accuracies obtained under closed-set and cross-seed and fine-tune scenarios. Column $2$ represents the net accuracy obtained for closed-set data which includes real and fake data from all four $GAN$ models. Columns $3$ to $6$ represent accuracies obtained for individual $GAN$ generated images. Column $7$ represents the net accuracy obtained for fine-tuned GAN data. The results highlight the generalization capability and efficiency of $GDA$-$Net$ in correctly attributing the $GAN$ generated images.}
 \vspace{1em}
\begin{tabular}{|c|c|c|c|c|c|c|}
\hline
\textbf{Method}        & \textbf{Closed-Set} & \textbf{\begin{tabular}[c]{@{}c@{}}Cross-Seed\\ ProGAN\end{tabular}} & \textbf{\begin{tabular}[c]{@{}c@{}}Cross-Seed\\ SNGAN\end{tabular}} & \textbf{\begin{tabular}[c]{@{}c@{}}Cross-Seed\\ DCGAN\end{tabular}} & \textbf{\begin{tabular}[c]{@{}c@{}}Cross-Seed\\ WGAN\end{tabular}}   & \textbf{\begin{tabular}[c]{@{}c@{}}Fine\\ tune\end{tabular}}\\  \hline
                                                               
\textbf{LeveFreq}      & 99.50                 & 83.50                                                                & 15.78                                                                 & 38.56                                                                  & 76.45                                                   & 33.38       \\ \hline
\textbf{AttNet}        & 98.88                & 89.44                                                              & 17.83                                                                 & 14.69                                                                  & 92.46                                                    & 47.50                \\ \hline
\textbf{DNA-Det}       & 100.00              & 95.05                                                                & 05.26                                                                & 69.14                                                                  & 94.12                                                    &  78.01         \\ \hline
\textbf{$GDA$-$Net$} & 100.00              & 99.90                                                                & 97.13                                                               & 99.96                                                               & 99.55                                                        & 98.57      \\ \hline
\end{tabular}

\label{comp}
\end{table}

\section{Conclusion}

In this paper, we proposed a generalized deepfake attribution network ($GDA$-$Net$) for attributing the $GAN$ generated images to its original architecture. The main goal of our method is to find the traces from the generated images that are $GAN$ architecture-specific. To achieve this goal, we have introduced a Feature extraction network that can extract architectural-level traces from the generated images using supervised contrastive learning. To further ensure that $GDA$-$Net$ focuses on the model architecture traces, we have added a denoising autoencoder such that $FEN$ will receive a feature map with the least content-dependent traces. To show the generalization of $GDA$-$Net$, we have used four different $GAN$s of $DCGAN$, $WGAN$, $ProGAN$ and $SNGAN$ and show that our method can correctly attribute the generated images. We have also compared our method with the existing attribution networks to highlight the effectiveness of $GDA$-$Net$.

\bibliographystyle{splncs04}
\bibliography{main}
\end{document}